\documentclass[authoryear,preprint,12pt]{elsarticle}

\usepackage[nopatch=footnote]{microtype}
\usepackage[T1]{fontenc}
\usepackage{lmodern}
\usepackage{amssymb}
\usepackage{amsmath}
\usepackage{amsopn} 
\usepackage{amsthm}

\usepackage{natbib}
\usepackage{hyperref}
\usepackage[ruled,vlined]{algorithm2e}
\usepackage{threeparttable}
\usepackage{booktabs}
\usepackage{multirow}
\usepackage{makecell}
\usepackage{xcolor}
\usepackage{subcaption}

\usepackage{setspace}

\begin{document}

\begin{frontmatter}

\title{Fast and Effective Redistricting Optimization via Composite-Move
Tabu Search}

\author{Hai Jin}
\author{Diansheng Guo\corref{cor1}} 
\ead{diansheng.guo@polyu.edu.hk}
\cortext[cor1]{Corresponding author}

\affiliation{organization={Department of Land Surveying and Geo-Informatics},
            addressline={The Hong Kong Polytechnic University, Kowloon},
            city={Hong Kong},country={China}}

\begin{abstract}
Spatial redistricting is a practical combinatorial optimization problem that demands high-quality solutions, rapid turnaround, and flexibility to accommodate multi-criteria objectives and interactive refinement. A central challenge is the contiguity constraint: enforcing contiguity in integer-programming or heuristic search can severely shrink the feasible neighborhood, weaken exploration, and trap the search in poor local optima. We introduce a \emph{composite-move Tabu search} (CM-Tabu) that systematically expands the feasible neighborhood space in Tabu search while preserving contiguity. When a boundary unit cannot be reassigned individually without disconnecting its district, our method identifies a minimal set of units that can move together, or a pair of units (or sets of units) that can be switched, as a contiguity-preserving composite move.  Candidate single-unit and composite moves are generated in linear time by analyzing each district's contiguity graph using articulation points and biconnected components. Extensive experiments demonstrate that the proposed approach substantially improves solution quality, run-to-run robustness, and computational efficiency relative to traditional Tabu search and other baselines. For example, in the Philadelphia case, the approach can consistently attain the theoretical global optimum in population-equality and support multi-criteria trade-offs.  CM-Tabu delivers optimization performance suitable for real-world practices and decision-support workflows.
\end{abstract}


\end{frontmatter}

\section{Introduction}

Many real-world decision-making problems rely on spatial combinatorial optimization, including political redistricting, school districting, service territory design, and facility location planning. In practice, an optimization method is rarely judged solely by whether it can improve an objective value on a benchmark. Instead, it must (i) reliably produce high-quality solutions, (ii) do so with fast turnaround suitable for iterative scenario testing, and (iii) remain flexible so that decision makers can incorporate multiple criteria and interactive refinement. Achieving these requirements simultaneously remains challenging for many spatial districting problems.

Political redistricting is a representative and societally important example. The task redraws legislative boundaries at various administrative levels---such as congressional, state legislative, or city council districts---while balancing multiple criteria including population equality, compactness, and preservation of political or community boundaries, under the hard constraint that each district must be geographically contiguous.

A wide range of computational approaches have been proposed, including clustering and region growing, location-allocation, space partitioning, graph partitioning, genetic algorithms, simulated annealing, Tabu search, and more recently, learning-based methods. A comprehensive review is provided in Section 2. While these methods can be effective in specific settings, many struggle to meet practical requirements simultaneously: some are fast but yield limited solution quality, some can optimize objectives but require extensive runtime or sensitive parameter tuning, and others provide little support for multi-criteria exploration or iterative human-guided adjustment.

A central challenge is the \textbf{contiguity constraint}. In integer-programming formulations, contiguity is difficult to encode without introducing large numbers of auxiliary variables and constraints, which severely limits scalability and practical applicability. In heuristic search (e.g., hill climbing, simulated annealing, Tabu search), contiguity is commonly handled in one of two ways: (i) allowing only moves that keep districts contiguous, or (ii) generating candidate solutions freely and filtering out those that violate contiguity. The former can drastically shrink the feasible neighborhood, while the latter can waste substantial computation exploring infeasible states; both weaken exploration and make it difficult to escape poor local optima.

This paper introduces \emph{Composite-Move Tabu search (CM-Tabu)}, a fast and effective redistricting optimizer that addresses this bottleneck by systematically expanding the feasible neighborhood space while preserving contiguity throughout the search. The key idea is to go beyond single-unit reassignment or simple pair swaps. When a boundary unit cannot be moved individually without disconnecting its district, CM-Tabu identifies a minimal set of units that can move together as a contiguity-preserving composite move; similarly, when beneficial, it considers switches that may involve sets of units while maintaining contiguity. Single-unit and composite moves are generated efficiently in linear time by analyzing each district's contiguity graph using articulation points (cut points) and biconnected components. As a result, CM-Tabu expands the feasible neighborhood at each iteration without compromising feasibility or efficiency, substantially improving the ability to escape local optima.

We evaluate CM-Tabu through extensive experiments, including a large-run study on the Iowa congressional problem and a multi-criteria Philadelphia City Council case study, demonstrating substantial improvements in solution quality, robustness, and runtime efficiency compared with traditional Tabu search and other baselines. For example, in the Iowa congressional case study (grouping 99 counties into 5 districts), 1,000-run experiments show that our approach (CM-Tabu) improves both solution quality and reliability: the median population deviation decreases by 84\% (from 2,627 to 371) and the interquartile range shrinks by 93\% (from 2,890 to 192), while maintaining sub-second runtime per run. Beyond redistricting, this contiguity-preserving neighborhood expansion approach can be extended to a broader class of districting and spatial partitioning problems where connectivity is required, such as school districting, service territory design, and facility location planning.

The contributions of this work include: (1) a contiguity-preserving neighborhood expansion based on minimal composite moves and composite switches that substantially enlarges the feasible move set under hard contiguity constraints; (2) a linear-time move-generation algorithm using articulation points and biconnected components, enabling practical use at scale; and (3) an integrated CM-Tabu optimizer and empirical evaluation showing fast, effective, and robust performance on real redistricting case studies.

The remainder of the paper is organized as follows. Section 2 reviews related work. Section 3 presents CM-Tabu, including contiguity-preserving move generation, objective evaluation, and the Tabu search procedure. Section 4 reports experimental results, and Section 5 concludes with discussion.

\section{Related Work}

Automated redistricting algorithms can be broadly classified into \emph{bottom-up agglomerative methods}, \emph{top-down divisive methods}, \emph{heuristic-based search methods, sampling-based methods, and learning-based methods}. These categories differ in how they represent space, handle constraints (especially contiguity), and balance optimization quality against runtime and practical usability.

\subsection{Bottom-up agglomerative methods}

Bottom-up agglomerative methods build districts by aggregating ``similar'' spatial units into regions under a contiguity constraint, often through hierarchical merging until reaching the desired number of districts. This group includes classical location-allocation methods \citep{hessNonpartisanPoliticalRedistricting1965,kalcsicsUnifiedTerritorialDesign2005,rios-mercadoLocationallocationimprovementHeuristicDistricting2021}, multi-kernel growth techniques \citep{vickreyPreventionGerrymandering1961,gearhartLegislativeDistrictingComputer1969,openshawOptimalZoningSystems1977}, and regionalization methods \citep{guoRegionalizationDynamicallyConstrained2008,guoDetectingSpatialCommunity2018}. Agglomerative methods are typically fast and naturally maintain contiguity by expanding districts through adjacent units. However, because key criteria are difficult to optimize directly during incremental growth, the resulting plans often require subsequent improvement by local or metaheuristic search.

\subsection{Top-down divisive and optimization-based methods}

Top-down divisive methods begin with the whole region and partition it into the target number of districts. Integer programming is a representative top-down formulation because it optimizes a global objective over a set of decision variables. In practice, however, expressing core redistricting constraints---especially \emph{geographic contiguity}---often requires many auxiliary variables and constraints \citep{ligmann-zielinskaSpatialOptimizationGenerative2008,dugosijaNewIntegerLinear2020,validiImposingContiguityConstraints2022,shahmizadPoliticalDistrictingMinimize2025}, which limits scalability and practical applicability \citep{covaExploratorySpatialOptimization2000}. As a result, integer linear programming (ILP) is not commonly used when strict contiguity must be guaranteed \citep{altmanPromisePerilsComputers2010}. Beyond ILP, additional optimization-based formulations include balanced centroidal power diagrams \citep{cohen-addadBalancedCentroidalPower2018}, multi-objective spanning tree models \citep{kimMultiobjectiveSpanningTree2018}, center-based modeling approaches \citep{kongCenterbasedModelingApproach2019}, hybrid algorithms for equal districting \citep{kongHybridAlgorithmEqual2021}, fairness-optimized column generation methods \citep{gurneeFairmanderingColumnGeneration2021}, and scalable multilevel multi-objective optimization frameworks \citep{swamyMultiobjectiveOptimizationPolitically2023,swamyPracticalOptimizationFramework2024}.

\subsection{Heuristic-based search and metaheuristics}

A large body of work focuses on heuristic-based search methods that start from an initial plan (often generated quickly via random growth) and iteratively improve it through local modifications such as moving units between neighboring districts. Representative approaches include local greedy search or hill climbing \citep{nagelSimplifiedBipartisanComputer1965}, simulated annealing \citep{browdySimulatedAnnealingImproved1990,gutierrez-andradeSimulatedAnnealingArtificial2019}, Tabu search \citep{bozkayaTabuSearchHeuristic2003}, evolutionary and genetic algorithms \citep{bergeySimulatedAnnealingGenetic2003,xiaoUnifiedConceptualFramework2008,tomczykEvolutionaryAlgorithmsSolving2024}, and dynamically weighted Voronoi diagrams \citep{riccaWeightedVoronoiRegion2008}.

Trajectory-based methods (e.g., greedy, simulated annealing, and Tabu) differ in how they accept non-improving moves and how they avoid cycling, which affects both runtime and solution quality. In redistricting, Tabu search is often effective because it can continue moving through non-improving steps while discouraging immediate reversals through a tabu list and restart strategy. Evolutionary approaches operate on populations of plans but typically require additional procedures to enforce contiguity during mutation and crossover \citep{xiaoUnifiedConceptualFramework2008}, and redistricting's strict equality objectives can make feasible recombination especially challenging.

\subsection{Sampling-based ensembles and learning-based approaches}

Sampling-based approaches (notably Markov chain Monte Carlo and related methods) are widely used to generate large ensembles of feasible plans for statistical analysis of fairness and bias \citep{defordRecombinationFamilyMarkov2020,fifieldAutomatedRedistrictingSimulation2020,mccartanSequentialMonteCarlo2023,dobbsRedistrictingOptimizationRecombination2023}. These methods emphasize probabilistic exploration of the feasible plan space rather than direct objective optimization \citep{kuengFairRedistrictingHard2019}. More recently, learning-based and decision-aware optimization approaches have been proposed to incorporate fairness objectives and policy considerations into model design \citep{levinAutomatedCongressionalRedistricting2019,choHumancenteredRedistrictingAutomation2020,koAllPoliticsLocal2022,chenExploringTradeoffsAutomated2023,ahmedDistrictNetDecisionawareLearning2024}. Such approaches often depend on high-quality training data or large plan ensembles that can be expensive to generate, and models trained in one geographic or institutional context can be difficult to transfer directly to others due to differences in spatial structure, demographics, and legal constraints.

\subsection{Contiguity constraint and neighborhood bottleneck}

Across these families of methods, spatial contiguity is typically handled in one of two ways. One is to encourage contiguity indirectly through objective design (e.g., distance-based penalties), which cannot guarantee contiguity in the final plan. The other is to enforce contiguity throughout the optimization process. For integer programming, enforcing contiguity generally requires many additional variables and constraints, making it difficult and inefficient at scale. For trajectory-based methods such as greedy search or Tabu search, contiguity is commonly enforced by permitting only those moves that do not disconnect districts. While this strategy guarantees feasibility, it can dramatically reduce the number of allowable moves, weaken exploration, and make it difficult to escape local optima.

This bottleneck motivates the approach developed in this paper: rather than relying exclusively on single-unit reassignment or simple swaps, we expand the feasible search neighborhood under strict contiguity by introducing contiguity-preserving composite moves, enabling trajectory-based metaheuristics to explore a much larger feasible space and achieve better optimization results without sacrificing feasibility or efficiency.

\section{Optimization under Contiguity Constraint: A New Approach}

We present a systematic and efficient approach for optimization problems with a hard contiguity constraint. Many metaheuristics---such as hill climbing, simulated annealing, and Tabu search---operate by iteratively modifying a current solution through local changes (moves), producing a trajectory of solutions that gradually improves objective quality. Under contiguity constraints, however, the set of feasible local changes can become extremely limited: a large fraction of boundary units cannot be moved individually without disconnecting the source district, and many seemingly reasonable swaps also violate contiguity. This dramatically reduces the feasible search neighborhood and weakens the search's ability to escape local optima.

Our approach addresses this bottleneck by explicitly analyzing within-district connectivity structure and expanding the feasible neighborhood while preserving contiguity throughout the search. At each iteration, we identify all candidate moves available under the current plan, including both \emph{traditional single-unit moves} and \emph{new composite (multi-unit) moves}. When a boundary unit cannot be reassigned individually due to contiguity, we construct a minimal set of units that can move together such that contiguity is maintained in both the source and destination districts after the move. Intuitively, these composite moves ``unlock'' boundary adjustments that are infeasible under single-unit-only search, substantially enlarging the feasible neighborhood at each step.

The same contiguity-aware framework also supports \emph{switch operations}. Traditional trajectory-based methods often consider pairwise swaps to improve strict criteria such as population equality. In our setting, a switch may also involve two composite moves (each potentially consisting of multiple units), enabling contiguity-preserving exchanges that are not reachable through single-unit swaps alone.

The neighborhood-expansion mechanism described above can be integrated with multiple trajectory-based optimizers. In principle, it can strengthen any method whose core operation is ``enumerate feasible local changes → select the best move → update the plan.'' In our experiments, combining this mechanism with \emph{Tabu search} yields the strongest overall performance, producing consistently higher-quality solutions with strong run-to-run reliability. We therefore present the integrated method as \textbf{Composite-Move Tabu search (CM-Tabu)}.

In addition to improved optimization power, the proposed approach is designed for practical efficiency. Candidate single-unit and composite moves are generated in linear time by analyzing each district's contiguity graph using articulation points (cut points) and biconnected components, allowing fast move enumeration and efficient scoring.  Because CM-Tabu is efficient, it can also be repeated from different random initial plans to generate a collection of high-quality alternative solutions, supporting interactive evaluation in decision-support settings. \emph{Algorithm~\ref{alg:overall}} summarizes the overall workflow.

\begin{algorithm}[!htbp]
    \caption{Overall Optimization Workflow of Our Approach (CM-Tabu)}
    \label{alg:overall}
    \SetKwBlock{RepeatSteps}{repeat a--c steps until a stopping condition is met}{}
    
    \textbf{1. Initialization:} Randomly generate an initial contiguous plan (Algorithm~\ref{alg:initial})\;
    
    \textbf{2. Optimization:} \RepeatSteps{
        \textbf{a.} Enumerate all feasible candidate moves (single-unit and composite) (Algorithm~\ref{alg:composite_move})\;
        \textbf{b.} Select the best move or switch, according to an objective function $f$\;
        \textbf{c.} Apply the move/switch; update Tabu memory and record the best plan found\;
    }
    
    \textbf{3. Output:} Return the best plan encountered during the search\;
    
    \textbf{4. Repetition (optional):} Repeat steps 1--3 to produce alternative high-quality plans.
\end{algorithm}

The remainder of Section 3 details each component of CM-Tabu: initialization under contiguity (Section 3.1), candidate move generation including composite moves and valid switches (Section 3.2), objective functions and efficient evaluation (Section 3.3), and the full CM-Tabu optimization procedure (Section 3.4).

\subsection{Initialization under Contiguity Constraint}

We generate an initial redistricting plan using a simple \emph{random seed-growing} procedure that guarantees contiguity. The input is a set of spatial units (for example, counties, precincts, or wards) together with their contiguity relationships (who touches whom). The goal is to construct a plan with a specified number of districts such that every unit is assigned to exactly one district, and every district is geographically contiguous.

The procedure works as follows. First, it randomly selects the required number of units as \emph{seeds}, and each seed becomes the starting unit of one district. Then, districts grow one unit at a time. During growth, each district repeatedly adds one unassigned neighboring unit (chosen at random) from its current boundary. This initialization method is intentionally randomized and does not attempt to optimize population equality, compactness, or any other criterion. This is appropriate for CM-Tabu because the optimization phase will improve the objective values. In addition, random initialization makes it easy to run the optimizer multiple times from different starting plans, which improves robustness and produces diverse high-quality alternatives for interactive decision support.

\begin{algorithm}[H]
    \caption{Initialization (seed-growing under contiguity)}
    \label{alg:initial}
    
    \SetKwInOut{Input}{Inputs}

    \Input{
        $S$: set of spatial units (there are $n$ units in total)\;
        $C$: contiguity structure (for example, an adjacency list or adjacency matrix)\;
        $r$: number of districts ($1 < r \ll n$)\;
    }
    \KwOut{An initial plan consisting of $r$ contiguous districts that collectively cover all units}

    \BlankLine
    Randomly select $r$ different units from $S$, each representing a district\;

    \While{all units are not yet assigned}{
        \ForEach{district (in turn)}{
            \If{district has unassigned neighboring units}{
                Randomly select one of the district's unassigned neighboring units and add it to the district\;
            }
        }
    }

    Output the resulting plan as the initial plan\;
\end{algorithm}

\subsection{Candidate Moves under Contiguity Constraint}

Given an initial plan with \emph{r} districts, we use Tabu search to optimize an objective function. Tabu search is a trajectory-based method that iteratively reassigns units between neighboring districts to seek improved solutions. We present the objective function in Section 3.3 and the full Tabu procedure in Section 3.4. Here we focus on how the contiguity constraint creates major challenges for optimization and how our method addresses them.

\begin{figure}[!htbp]
    \centering
    \includegraphics[width=0.8\textwidth]{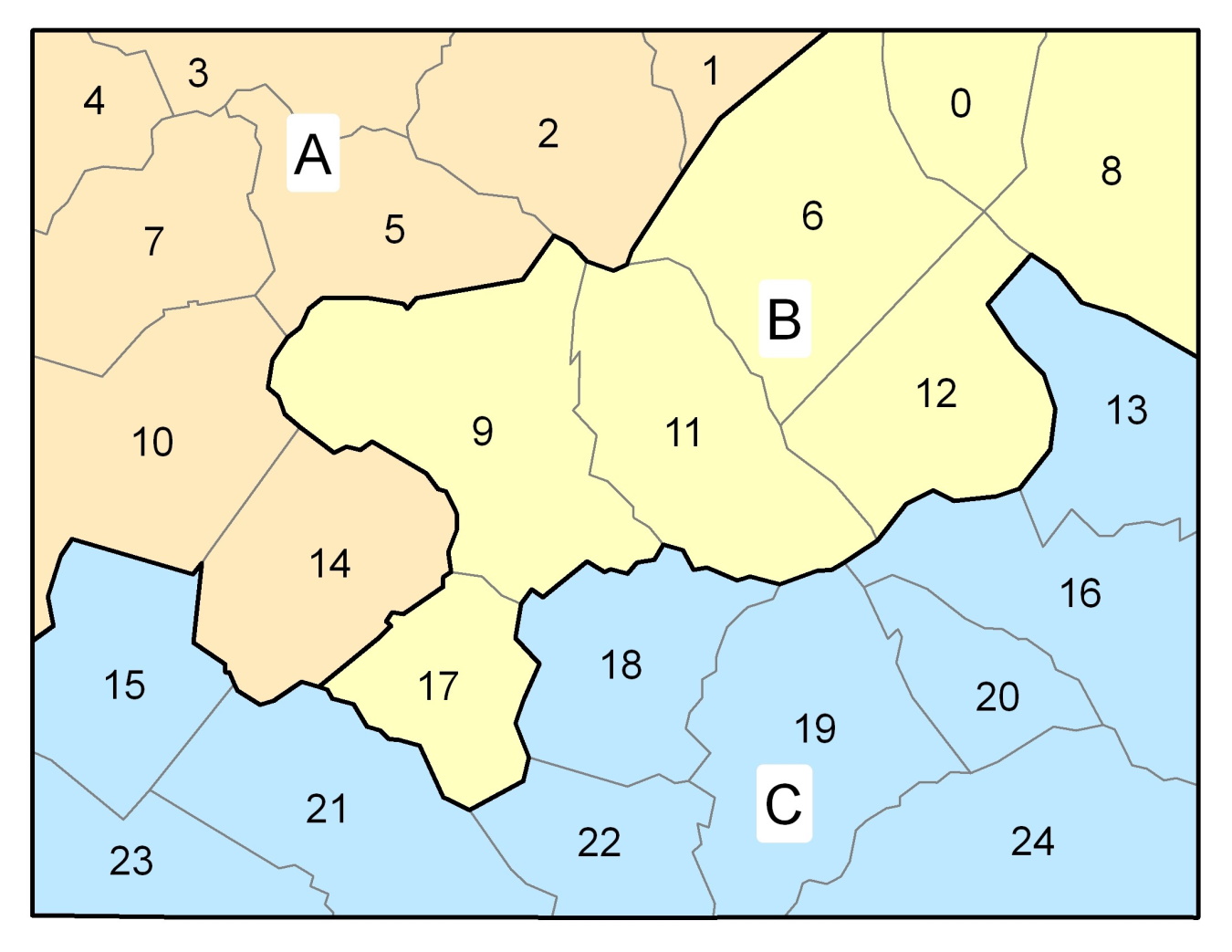}
    \caption{An example data set to demonstrate candidate moves under contiguity constraint. See Table~\ref{tab:candidate_moves} for the candidate moves with traditional methods and the new method.}
    \label{fig:candidate}
\end{figure}

Most existing trajectory-based methods consider only single-unit moves (moving one unit at a time) or pairwise switches (swapping two units) at each iteration. Under a hard contiguity constraint, however, the feasible neighborhood defined by such moves can be extremely limited.  Figure~\ref{fig:candidate} illustrates this issue using a simple example of 25 spatial units grouped into three districts: A, B, and C. Between districts A and B, for example, only three units (1, 5, 14) can move from A to B and only two units (6, 17) can move from B to A without breaking contiguity.  Among these five units, only four pairs can be switched (1 with 17; 5 with 6; 5 with 17; and 14 with 6). Therefore, traditional methods provide only nine feasible moves (single-unit moves or pair switches) between districts A and B. By contrast, our approach identifies nineteen feasible moves between A and B, including single-unit moves, contiguity-preserving composite moves, and valid pair switches (Figure~\ref{fig:candidate} and Table~\ref{tab:candidate_moves}). This doubles the number of feasible options per iteration; over T iterations, that multiplicative gain compounds into an exponential increase in the number of reachable move sequences (roughly two to the power of T).

\begin{table}[!htbp]
    \centering
    \begin{threeparttable}
        \caption{Candidate moves for the data shown in Figure~\ref{fig:candidate}.}
        \label{tab:candidate_moves}
        \scriptsize
        \begin{tabular}{lrcrc}
            \toprule
            & \multicolumn{2}{c}{\textbf{Traditional Methods}} & \multicolumn{2}{c}{\textbf{Our Approach}} \\ 
            \cmidrule(lr){2-3} \cmidrule(lr){4-5}
            \textbf{Direction} & \textbf{Single-Unit Moves} & \makecell{\textbf{Switch} \\ \textbf{Moves}} & \makecell{\textbf{Single-Unit and Composite Moves} \\ \textbf{(Cut points are underlined)}} & \makecell{\textbf{Switch} \\ \textbf{Moves}} \\ 
            \midrule
            $A \rightarrow B$ & $\{1\}, \{5\}, \{14\}$ & \multirow{2}{*}{\makecell{4 switch \\ moves}} & $\{1\}, \{5\}, \{14\}, \{\underline{2}, 1\}, \{\underline{10}, 14\}$ & \multirow{2}{*}{\makecell{10 switch \\ moves}} \\ 
            \cmidrule(lr){1-1} \cmidrule(lr){2-2} \cmidrule(lr){4-4}
            $A \leftarrow B$ & $\{6\}, \{17\}$ & & $\{6\}, \{17\}, \{\underline{11}, 9, 17\}, \{\underline{9}, 17\}$ & \\ 
            \midrule
            $A \rightarrow C$ & $\{14\}$ & \multirow{2}{*}{\makecell{1 switch \\ move}} & $\{14\}, \{\underline{10}, 14\}$ & \multirow{2}{*}{\makecell{1 switch \\ move}} \\ 
            \cmidrule(lr){1-1} \cmidrule(lr){2-2} \cmidrule(lr){4-4}
            $A \leftarrow C$ & $\{15\}$ & & $\{15\}, \{\underline{21}, 15, 23\}$ & \\ 
            \midrule
            $B \rightarrow C$ & $\{8\}, \{12\}, \{17\}$ & \multirow{2}{*}{\makecell{6 switch \\ moves}} & $\{8\}, \{12\}, \{17\}, \{\underline{11}, 9, 17\}, \{\underline{9}, 17\}$ & \multirow{2}{*}{\makecell{18 switch \\ moves}} \\ 
            \cmidrule(lr){1-1} \cmidrule(lr){2-2} \cmidrule(lr){4-4}
            $B \leftarrow C$ & $\{13\}, \{18\}$ & & \makecell[r]{$\{13\}, \{18\}, \{\underline{16}, 13\}, \{\underline{21}, 15, 23\}$, \\ $\{\underline{19}, 13, 16, 20, 24\}, \{\underline{22}, 15, 21, 23\}$} & \\ 
            \midrule
            \textbf{Total} & \multicolumn{2}{c}{\textbf{23 moves}} & \multicolumn{2}{c}{\textbf{53 moves}} \\ 
            \bottomrule
        \end{tabular}
    \end{threeparttable}
\end{table}

\subsubsection{Single-Unit Moves and Composite Moves}

To expand the search space under a hard contiguity constraint, our approach systematically analyzes within-district adjacency and constructs \textbf{composite moves} for boundary units that cannot move independently.

Specifically, if a boundary unit \emph{u} lies between two districts but cannot be reassigned by itself without breaking contiguity, our method identifies a \emph{minimal set of units} (including \emph{u}) that can be moved together so that contiguity is preserved in both the source and destination districts after the move. For example, in Figure~\ref{fig:candidate}, unit 10 in district A cannot move to district B as a single-unit move, but it can move together with unit 14 as a composite move. Similarly, eight units in Figure~\ref{fig:candidate} (2 and 10 in A; 9 and 11 in B; and 16, 19, 21, and 22 in C) cannot move under traditional single-unit or simple-swap methods, yet each becomes movable through a composite move in our approach. Table~\ref{tab:candidate_moves} summarizes the single-unit and composite moves produced by our method.

This neighborhood expansion has an immediate practical impact. For the example in Figure~\ref{fig:candidate}, traditional Tabu search yields 23 feasible moves (single-unit moves plus valid pair switch moves), while our method yields 53 (Table~\ref{tab:candidate_moves}). Because the neighborhood is evaluated repeatedly over many iterations, expanding the feasible move set at each iteration substantially improves the ability to escape poor local optima and increases run-to-run reliability, as demonstrated in Section 4.

\subsubsection{Efficient Identification of Composite Moves}

\begin{figure}[!htbp]
    \centering
    \includegraphics[width=0.9\textwidth]{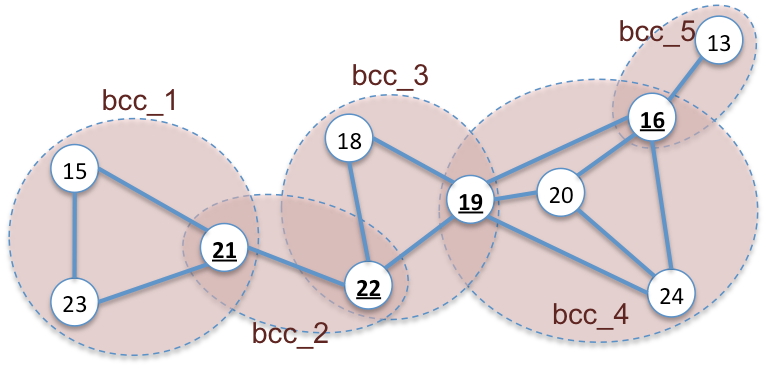}
    \caption{The contiguity relationship among the spatial objects in the district C in Figure~\ref{fig:candidate}. Neighbors are connected with edges, cut points are underlined, and dash-line ellipses show five biconnected components (bcc).}
    \label{fig:contiguity}
\end{figure}

We now present an efficient algorithm to generate all candidate moves (single-unit and composite) in linear time. We represent the contiguity relationships among units within a district as a graph, where each unit is a node and an edge connects two adjacent units (Figure~\ref{fig:contiguity}). If removing a unit \emph{u} from the district's contiguity graph splits the graph into two or more disconnected components, then \emph{u} is an \emph{articulation point} (also called a cut point). A \emph{biconnected component} is a maximal subgraph that cannot be disconnected by removing any single node \citep{gabowPathbasedDepthfirstSearch2000}. Figure~\ref{fig:contiguity} shows the contiguity graph of district C in Figure~\ref{fig:candidate}, which contains four cut points and five biconnected components.

\begin{figure}[!htbp]
    \centering
    \includegraphics[width=0.9\textwidth]{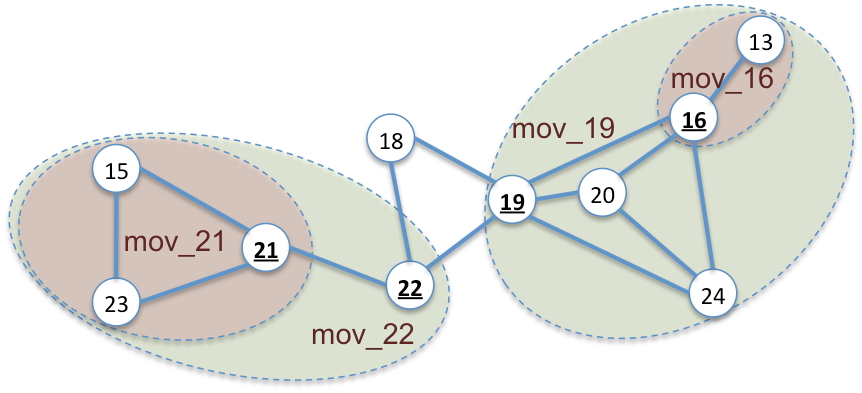}
    \caption{Composite moves (i.e., multi-object moves) for cut points. Each cut point will move as a composite move to maintain contiguity. For example, object 22 will move together with objects 15, 21, and 23 so that the remaining graph is still contiguous.}
    \label{fig:composite_move}
\end{figure}

The algorithm proceeds in two stages (Algorithm~\ref{alg:composite_move}). \emph{First}, it finds all cut points and biconnected components in each district using a depth-first search (DFS) method \citep{gabowPathbasedDepthfirstSearch2000,tarjanDepthfirstSearchLinear1972}. The complexity of the DFS algorithm is \(O(n)\), where \emph{n} is the number of spatial units in each district.

\begin{algorithm}[H]
    \small
    \setstretch{0.95}
    \caption{Identifying Composite Moves}
    \label{alg:composite_move}
    \SetKwInOut{Input}{Inputs}
    \SetKwInOut{Output}{Outputs}

    \Input{
        $S_d$: the set of spatial objects in a district $d$;\\
        $C_d$: a contiguity matrix of the objects in $S_d$;\\
        $A_d$: attribute vector for each object in $S_d$;
    }
    \Output{CompositeMoves: a list of composite moves for district $d$}

    Initialize CompositeMoves = $\emptyset$ and LeafBCC = $\emptyset$\;

    Find all cut points and biconnected components using DFS ($S_d, C_d$) \citep{gabowPathbasedDepthfirstSearch2000}\;
    \Indp
    $bcc.CPT$: the set of cut points that a biconnected component $bcc$ contains\;
    $cpt.BCC$: the set of biconnected components that a cut point $cpt$ belongs to\;
    $cpt.maxC = \emptyset$, which will keep the largest component for $cpt$\;
    $cpt.restC = \emptyset$, which will keep the union of other components of $cpt$\;
    \Indm

    \ForEach{biconnected component $bcc$}{
        \If{$|bcc.CPT| = 1$}{
            add $bcc$ to LeafBCC\;
        }
    }

    \While{LeafBCC is not empty}{
        $bcc = \text{next biconnected component in LeafBCC}$\;
        \If{$bcc.CPT$ is not empty}{
            Let $cpt = \text{the only cut point in } bcc.CPT$\;
            \eIf{$\text{size}(bcc) > \text{size}(cpt.maxC)$}{
                $cpt.restC = cpt.restC \cup cpt.maxC$\;
                $cpt.maxC = bcc$\;
            }{
                $cpt.restC = cpt.restC \cup bcc$\;
            }
            Remove $bcc$ from $cpt.BCC$\;
            \If{$|cpt.BCC| = 1$ and $\text{size}(cpt.maxC) < \text{size}(S_d) - \text{size}(cpt.maxC \cup cpt.restC) + 1$}{
                $cpt.restC = cpt.maxC \cup cpt.restC$\;
                $bccR = \text{the only remaining biconnected component in } cpt.BCC$\;
                Let $cpt.BCC = \emptyset$ and remove $cpt$ from $bccR.CPT$\;
                \If{now $|bccR.CPT| = 1$}{
                    add $bccR$ to LeafBCC\;
                }
            }
            \If{$cpt.BCC = \emptyset$}{
                Aggregate the attribute vectors $A_d$ over all units in $cpt.restC$ (including $cpt$ itself) to form the attributes of composite move $cpt$\;
                Add $cpt$ to \textbf{CompositeMoves} as a new composite move\;
            }
        }
    }
\end{algorithm}

\emph{Second}, for each cut point, the algorithm constructs one composite move, as illustrated in Figure~\ref{fig:composite_move}. By definition, biconnected components are connected only through cut points. If we treat each biconnected component as a single ``block'', the resulting block--cut graph forms a tree (Figure~\ref{fig:contiguity}). A biconnected component is a leaf in this tree if it connects to only one cut point (for example, bcc\_1 and bcc\_5 in Figure~\ref{fig:contiguity}). Since removing a cut point can split the district into multiple components, our strategy is to keep the largest resulting component as the main part of the district and combine the cut point with the remaining smaller components to form a composite move. The size of a component can be measured by the number of units it contains or by other quantities such as total population; here we use the number of units.

The algorithm starts from leaf biconnected components and traverses the tree bottom-up to determine composite moves. During this traversal, attribute values within each composite move are aggregated so that each composite move can be evaluated efficiently without scanning all units in it.

The time complexity of Algorithm~\ref{alg:composite_move} is \(O(n)\), where \emph{n} is the number of units in the district. Algorithm~\ref{alg:composite_move} is applied to each district. Each non-cut point produces a feasible single-unit move, and each cut point produces a composite move. Among these moves, those on the border between two districts are considered candidate moves between those neighboring districts. The same unit (for example, unit 14 in district A) may have feasible moves to different neighboring districts (for example, B or C), which are treated as different candidate moves. The list of candidate moves is updated after each accepted move (Step 2 in Algorithm~\ref{alg:overall}).

\subsubsection{Switch Moves: Exchanging Two Candidate Moves}

Pair switches (exchanging two candidate moves between two neighboring districts) are often useful for improving certain criteria, such as population equality \citep{nagelSimplifiedBipartisanComputer1965,bozkayaTabuSearchHeuristic2003}. Compared with existing methods that consider pairwise switches, our switching operation is more general because each side of a switch may involve more than one unit. For example, in Figure~\ref{fig:candidate}, the composite move \{2, 1\} and the composite move \{9, 17\} can be switched between their districts while preserving contiguity.

However, not all pairs of feasible moves can be switched while maintaining contiguity. For example, in Figure~\ref{fig:candidate}, unit 14 and unit 17 cannot be switched even though each can move individually. We address this by applying the validity check. Let \(M_1\) and \(M_2\) be two candidate moves, \(B_1\) and \(B_2\) be the boundary contacts that \(M_1\) and \(M_2\) share with their respective destination districts. Let \(B_s\) be the shared boundary between \(M_1\) and \(M_2\). If \(B_1 \subseteq B_s\) or \(B_2 \subseteq B_s\), then we cannot switch the two moves. Table~\ref{tab:candidate_moves} shows the total number of valid pair switches (i.e., switch moves) between neighboring districts.

\subsection{Objective Function (or Optimization Criteria)}

Different optimization problems can use very different objective functions (that is, optimization criteria). Because this paper uses political redistricting to present and evaluate our method, we focus on criteria commonly used in political redistricting. In addition to the contiguity constraint, \emph{population equality} is typically the most important criterion, as reflected in U.S. laws and state constitutions. In practice, additional criteria are often considered to limit gerrymandering and improve plan quality, although the specific set varies by application. Common examples include \emph{compactness of district shapes}, \emph{preservation of communities of interest}, and \emph{respect for existing political boundaries}. Some criteria, for example, ``communities of interest'', are inherently subjective and require user input and interpretation. In our previous work, we developed a visual analytics framework that supports interactive evaluation and iterative refinement, providing a way to incorporate subjective criteria into the optimization process \citep{guoIRedistrictGeovisualAnalytics2011}.

The focus of this paper is to introduce and evaluate the new optimization method. Therefore, for clear comparison and visual checking of quality, we choose two commonly used, well defined, and probably also the most challenging criteria in redistricting: population equality and compactness of district shapes.

\textbf{Population equality (PopDev)}

Population equality requires that district populations be as equal as possible to ensure the ``one person one vote'' principle. We measure population equality using ``population deviation'' (\(\mathit{PopDev}\)), defined as the sum, over all districts, of absolute differences between each district's actual population (\(p_i\)) and its ideal population, which is the total population (\(\mathit{Pop}\)) divided by the number of districts \(r\).

\begin{equation}
PopDev = \sum_{i = 1}^{r}\left\lfloor \left| p_{i} - \frac{Pop}{r} \right| \right\rfloor 
\end{equation}

\textbf{Compactness (Polsby--Popper)}

Many compactness measures have been proposed in the literature \citep{maceachrenCompactnessGeographicShape1985}. We use the Polsby-Popper index (\(\mathit{PPI}\)) \citep{polsbyPartisanGerrymanderingHarms1991}, which is commonly used in redistricting. \(\mathit{PPI}\) measures the compactness of a district \emph{P} as the ratio between the district area (\(\alpha\)) and the area of a circle that has the same perimeter (\(\rho\)) of \emph{P}. \(\mathit{PPI}\) values range between 0 and 1, with 1 corresponding to a perfect circle.

\begin{equation}
\mathit{PPI} = \frac{4\pi\alpha}{\rho^{2}}
\end{equation}

To combine compactness with population deviation in a single objective, we reverse and normalize \(\mathit{PPI}\) to form a derived compactness penalty for each district, and then sum across districts to obtain the plan-level compactness term. We use the normalization described in Equation~\ref{eq:compactness} so that the compactness term is on a comparable scale to the population term.

\begin{equation}
\mathit{Compactness} = \sum_{i = 1}^{r}{\frac{\mathit{Pop}}{1000}\left( 1 - {\mathit{PPI}}_{i} \right)}
\label{eq:compactness}
\end{equation}

\textbf{Overall Objective}

The overall objective function \emph{f} is a weighted sum of the selected criteria. In this paper, the objective includes \(\mathit{PopDev}\) and \(\mathit{Compactness}\), with user-specified weights. For other applications, the same framework can incorporate different criteria and weights as needed.

\subsubsection{Efficient Evaluation of Candidate Moves}

Based on a given objective function \emph{f}, each candidate move \emph{m} is assigned a score, \(\delta_m\), defined as the change in objective value caused by applying the move. In other words, \(\delta_m = f(P) - f(P_m)\), where \emph{P} is the current plan and \(P_m\) is the new plan after making the move \emph{m}. The move with the largest score is the best move if the objective function is to be minimized.

To achieve high efficiency, we calculate the score for each move using aggregated attribute values of the move and aggregated information from the two involved districts. This strategy is known as ``dynamic scoring'' \citep{altmanBARDBetterAutomated2011}. For example, given two districts A and B and a set of candidate moves between them, we first compute aggregated attributes for each district (such as total population and dissolved shape boundary). We then compute the score of each candidate move using its aggregated attributes (such as population and dissolved shape boundary), which are obtained during candidate-move generation in Section 3.2. As a result, we can score all candidate moves and identify the best move in linear time.

\subsubsection{Efficient Evaluation of Pair Switches}

Evaluating pair switches can be time-consuming if we enumerate and score all possible pairs. Because pair switches are mainly used to improve population equality, we develop an efficient strategy to find the best switch without enumerating all pairs.

Suppose there are two neighboring districts, A and B, each with a set of candidate moves. To find the best pair to switch, we first order the moves in each district by the population they would transfer. Then, for a given move \emph{u} in A, using the population transfer of \emph{u} and the current populations of A and B, we compute the target population transfer of an ideal move in B to switch with \emph{u}. Since the moves in B are already ordered, we can use a binary search to locate the move \emph{m} in B whose population transfer is closest to the target value. We then examine a small number of moves on both sides of \emph{m} in the ordering and select the best partner move \emph{v} for \emph{u} according to the overall objective function. Note that this best partner \emph{v} is defined relative to \emph{u} only. Repeating this procedure for each move in A yields the best switch between A and B.

The time complexity for evaluating pair switches is \(O(n \log n)\), where \emph{n} is the total number of moves in A and B. Finally, we compare the best single move identified in Section 3.3.1 with the best switch identified here to determine the overall best move.

\subsection{Optimization}

\begin{figure}[!htbp]
    \centering
    \includegraphics[width=1\textwidth]{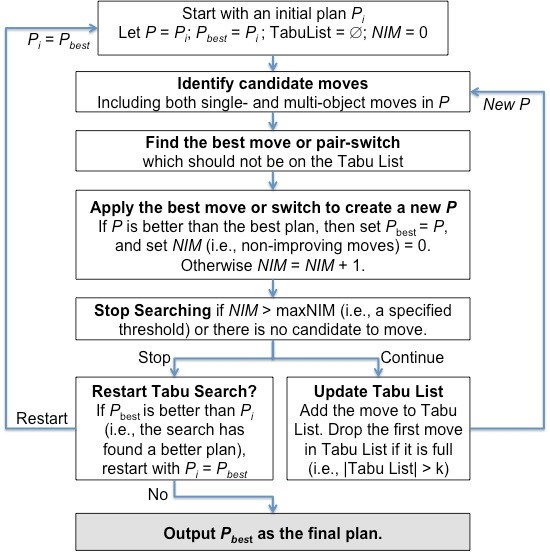}
    \caption{The Tabu search algorithm to optimize an initial redistricting plan.}
    \label{fig:tabu}
\end{figure}

So far, we have introduced an initialization method, an efficient algorithm for identifying candidate moves under the contiguity constraint, an objective function, and strategies for efficient move evaluation. We now present the Tabu search algorithm that integrates these components to efficiently and effectively derive high-quality redistricting plans. Tabu search has been used in many applications and has been shown to outperform alternative approaches \citep{gloverTabuSearch1997,battitiGreedyProhibitionReactive1999,bozkayaTabuSearchHeuristic2003}. Our Tabu search algorithm is shown in Figure~\ref{fig:tabu}.

The optimization procedure improves an initial plan by iteratively moving units from one district to a neighboring district. At each iteration, the algorithm evaluates all feasible single-unit moves and composite moves (including pair switches) and selects the best move according to the objective function. After the selected move is applied, the list of candidate moves is updated, and the next best move is identified. This process repeats until a stopping condition is met.

What makes Tabu search distinctive is its short-term memory mechanism for avoiding recently explored paths and encouraging the search to escape local optima. Specifically, the algorithm maintains a tabu list that records the most recent moves; moves on this list are temporarily prohibited until they expire from the list. The tabu list length, denoted by \emph{k} (the number of prohibited recent moves), is typically much smaller than the number of units \emph{n}. In our experiments, we set \(k = 0.08n\). Tabu search also allows non-improving moves, that is, the best available move at an iteration may not reduce the objective value. Allowing non-improving moves helps the search move out of local optima and potentially reach better solutions later. The search stops when the number of consecutive non-improving moves exceeds a threshold \(\mathit{maxNIM}\). In our experiments, we set \(\mathit{maxNIM} = 3n\).

By changing parameter settings, the algorithm in Figure~\ref{fig:tabu} can be converted to two other trajectory-based optimization methods: local greedy search (hill climbing) and the Kernighan--Lin algorithm. If \(k = 0\) (no tabu) and \(\mathit{maxNIM} = 0\) (does not allow non-improving moves), the algorithm becomes local greedy search, which is fast but often produce poor optimization quality. If \(k = \infty\) and \(\mathit{maxNIM} = \infty\) (i.e. each move can move once and the search stops when no valid move remains), the algorithm becomes equivalent to the Kernighan--Lin (K-L) procedure, originally developed for graph partitioning \citep{kernighanEfficientHeuristicProcedure1970} and widely used in network and graph optimization \citep{newmanModularityCommunityStructure2006}.

Moreover, by enabling or disabling our contiguity-preserving neighborhood expansion approach (which introduces composite moves), the algorithm in Figure~\ref{fig:tabu} can be configured into six different methods, summarized in Table~\ref{tab:optimization_methods}. Without composite moves, we have three traditional trajectory-based optimization methods: Greedy, K--L, and Tabu. With composite moves enabled, we obtain three enhanced variants: \(\mathrm{Greedy}^*\), \(\mathrm{K\text{--}L}^*\), and \(\mathrm{Tabu}^*\) (i.e., CM-Tabu), where the star (\({}^*\)) indicates support for composite moves. Our experiments in Section 4 show that each enhanced method substantially outperforms its traditional counterpart while by a large margin while remaining computationally efficient.

\begin{table}[!htbp]
    \centering
    \begin{threeparttable}
        \caption{Different optimization methods that are implemented and compared.}
        \label{tab:optimization_methods}
        \footnotesize
        \begin{tabular}{|l|c|l|l|l|}
            \hline
             & \makecell{\textit{Composite} \\ \textit{moves}} & \makecell{\textit{Tabu List} \\ \textit{Length (k)}} & \makecell{\textit{Maximum consecutive} \\ \textit{non-improving moves} \\ \textit{allowed}} & \textit{Time Complexity} \\
            \hline
            Greedy & No & $k = 0$ & $\mathit{maxNIM} = 0$ & $O(mn \log n), m \ll n$ \\
            \hline
            K--L & No & $k = \infty$ & $\mathit{maxNIM} = \infty$ & $O(n^2 \log n)$ \\
            \hline
            Tabu & No & $k \ll n$ & $\mathit{maxNIM} = 3n$ & $O(n^2 \log n)$ \\
            \hline
            $\mathrm{Greedy}^*$ & Yes & $k = 0$ & $\mathit{maxNIM} = 0$ & $O(mn \log n), m \ll n$ \\
            \hline
             $\mathrm{K\text{--}L}^*$  & Yes & $k = \infty$ & $\mathit{maxNIM} = \infty$ & $O(n^2 \log n)$ \\
            \hline
             $\mathrm{Tabu}^*$ (i.e., & \raisebox{-1.2ex}{Yes} & \raisebox{-1.2ex}{$k \ll n$} & \raisebox{-1.2ex}{$\mathit{maxNIM} = 3n$} & \raisebox{-1.2ex}{$O(n^2 \log n)$} \\
            CM-Tabu) & & & & \\
            \hline
        \end{tabular}
    \end{threeparttable}
\end{table}

We also implemented a genetic algorithm based on \citep{xiaoUnifiedConceptualFramework2008} for comparison. The genetic algorithm starts with a set of initial plans generated using the method in Section 3.1. Readers are referred to \citep{xiaoUnifiedConceptualFramework2008} for additional details. In our experiments, each generation contains 50 plans, and evolution continues for 200 generations. The best plan found during evolution is returned as the final output.

\section{Performance Evaluation}

\begin{figure}[!htbp]
    \centering
    \includegraphics[width=0.9\textwidth]{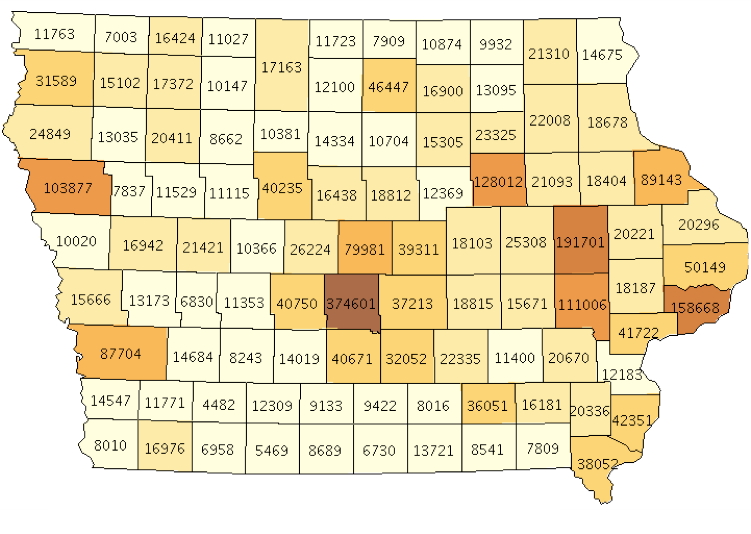}
    \caption{The population of 99 Iowa counties from the 2000 census data.}
    \label{fig:iowa_pop}
\end{figure}

We evaluate the proposed methods using Iowa congressional redistricting based on the 2000 Census as a case study. Iowa has 99 counties that must be divided into five congressional districts (Figure~\ref{fig:iowa_pop}). The total population is 2,926,324, so the ideal population per district is 585,265. In this research, two spatial units are considered contiguous if they share at least a segment of boundary.

\subsection{Optimizing Population Equality Only}

Our first experiment considers only the population-equality criterion, measured by population deviation (\(\mathit{PopDev}\)). We compare seven methods: the genetic algorithm, local greedy search (Greedy), Kernighan--Lin (K--L), Tabu search (Tabu), and their enhanced variants that enable composite moves (i.e., \(\mathrm{Greedy}^*\), \(\mathrm{K\text{--}L}^*\), and \(\mathrm{Tabu}^*\)). Each method is run 1,000 times to produce 1,000 plans that optimize \(\mathit{PopDev}\). Each run begins from a different random initial plan and therefore may produce a different final plan.

We compare the seven methods using the \(\mathit{PopDev}\) scores of their 1,000 final plans. Table~\ref{tab:performance_iowa} summarizes the results, including the minimum (best), the 5th percentile, the 25th percentile (first quartile, \(Q1\)), the median, the 75th percentile (third quartile, \(Q3\)), the 95th percentile, and the max (worst). To assess reliability, Table~\ref{tab:performance_iowa} also reports the interquartile range (\(\mathit{IQR} = Q3 - Q1\)) and the standard deviation (\(\mathit{StdDev}\)).

\begin{table}[!htbp]
    \centering
    \begin{threeparttable}
        \caption{Performance evaluation with Iowa Data.}
        \label{tab:performance_iowa}
        \footnotesize
        \begin{tabular}{|r|r|r|r|r|r|r|r|}
            \hline
            \multicolumn{1}{|c|}{\multirow{2}{*}{\textit{\begin{tabular}[c]{@{}c@{}}1000 Runs,\\ PopDev Only\end{tabular}}}} & \multicolumn{1}{c|}{\multirow{2}{*}{\textit{\begin{tabular}[c]{@{}c@{}}Genetic\\ Algorithm\end{tabular}}}} & \multicolumn{3}{c|}{\textit{\begin{tabular}[c]{@{}c@{}}Trajectory-based \\ optimization methods\end{tabular}}} & \multicolumn{3}{c|}{\makecell{\textit{Combined with} \\ \textit{our approach}}} \\ \cline{3-8} 
            \multicolumn{1}{|c|}{} & \multicolumn{1}{c|}{} & \textit{Greedy} & \textit{K--L} & \textit{Tabu} & \textit{Greedy}$^*$ & \textit{K--L}$^*$ & \textit{Tabu}$^*$ \\ \hline
            Min & 7,089 & 713 & 205 & 77 & 755 & 159 & 33 \\ \hline
            5\% & 33,133 & 2,945 & 807 & 223 & 1,939 & 459 & 167 \\ \hline
            $Q1$ (25\%) & 56,705 & 7,375 & 1,703 & 417 & 4,239 & 915 & 289 \\ \hline
            Median (50\%) & 74,159 & 13,461 & 2,627 & 629 & 7,523 & 1,401 & 371 \\ \hline
            $Q3$ (75\%) & 93,313 & 33,883 & 4,593 & 1,025 & 12,779 & 2,289 & 481 \\ \hline
            95\% & 120,914 & 343,865 & 10,079 & 3,571 & 33,193 & 4,735 & 775 \\ \hline
            Max & 198,058 & 1,271,207 & 84,301 & 62,707 & 617,469 & 20,837 & 5,997 \\ \hline
            $\mathit{IQR} (Q3 - Q1)$ & 36,608 & 26,508 & 2,890 & 608 & 8,540 & 1,374 & 192 \\ \hline
            $\mathit{StdDev}$ & 26,745 & 147,434 & 6,050 & 3,332 & 39,796 & 1,794 & 316 \\ \hline
            Time (s/run)\textsuperscript{a} & 1.8 & 0.002 & 0.02 & 0.05 & 0.002 & 0.02 & 0.08 \\ \hline
        \end{tabular}
        \begin{tablenotes}
            \footnotesize
            \item [a] We implemented both the traditional methods (Greedy, K--L, Tabu) and the enhanced methods (\(\mathrm{Greedy}^*\), \(\mathrm{K\text{--}L}^*\), \(\mathrm{Tabu}^*\)) using the same efficient strategies described in this paper. Therefore, the running times reported for the traditional methods are faster than some existing implementations in the literature or in packages such as BARD \citep{altmanBARDBetterAutomated2011}. The experiments were run on a MacBook Pro laptop with a 2.66 GHz Intel processor.
        \end{tablenotes}
    \end{threeparttable}
\end{table}

\begin{figure}[!htbp]
    \centering
    \includegraphics[width=1\textwidth]{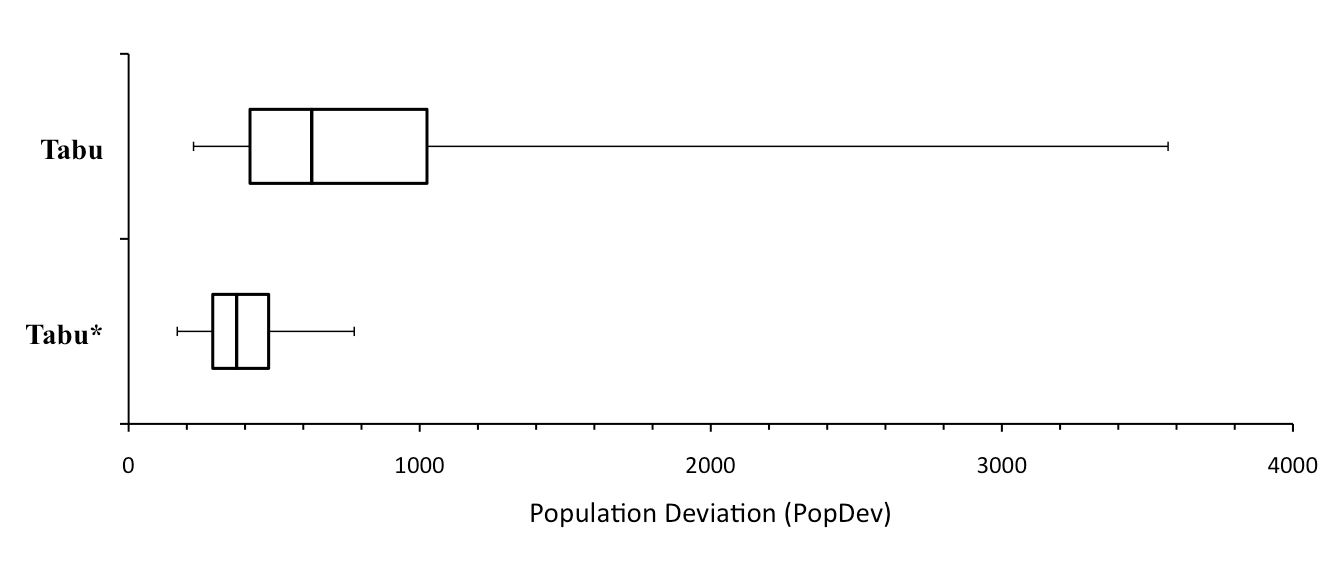}
    \caption{Comparing the performance of Tabu and \(\mathrm{Tabu}^*\) (see Table~\ref{tab:performance_iowa}). The whiskers show the 5th and 95th percentiles.}
    \label{fig:performance_iowa}
\end{figure}

Results show that enabling our contiguity-preserving neighborhood expansion (composite moves) yields substantial improvements: \(\mathrm{Greedy}^*\), \(\mathrm{K\text{--}L}^*\), and \(\mathrm{Tabu}^*\) consistently outperform their corresponding traditional versions by a large margin (\(p\text{-value} < 2 \times 10^{-16}\), Mann--Whitney--Wilcoxon test). Among all methods, \(\mathrm{Tabu}^*\) achieves the best overall performance in both optimization quality (lower \(\mathit{PopDev}\)) and reliability (smaller \(\mathit{IQR}\) and standard deviation), as illustrated in Figure~\ref{fig:performance_iowa}. Although the Iowa case involves only 99 spatial units, it is a challenging population-equality problem because county populations are large and severely limit how population can be redistributed while maintaining contiguity. In the Philadelphia case study (Section 4.2), \(\mathrm{Tabu}^*\) consistently produces plans that attain the theoretical global optimum for population equality (\(\mathit{PopDev} = 0\)), which is very difficult for existing methods to achieve \citep{gopalanPhiladelphiaDistrictingContest2013}.

\subsection{Optimizing Both Population Equality and Shape Compactness}

Our second experiment considers both population deviation (\(\mathit{PopDev}\)) and compactness. We use Philadelphia City Council redistricting based on the 2010 Census to demonstrate joint optimization of population equality and compactness. In this case, 1,687 ward divisions are partitioned into 10 districts. The total population is 1,526,006.

\begin{figure}[!htbp]
    \centering
    \includegraphics[width=1\textwidth]{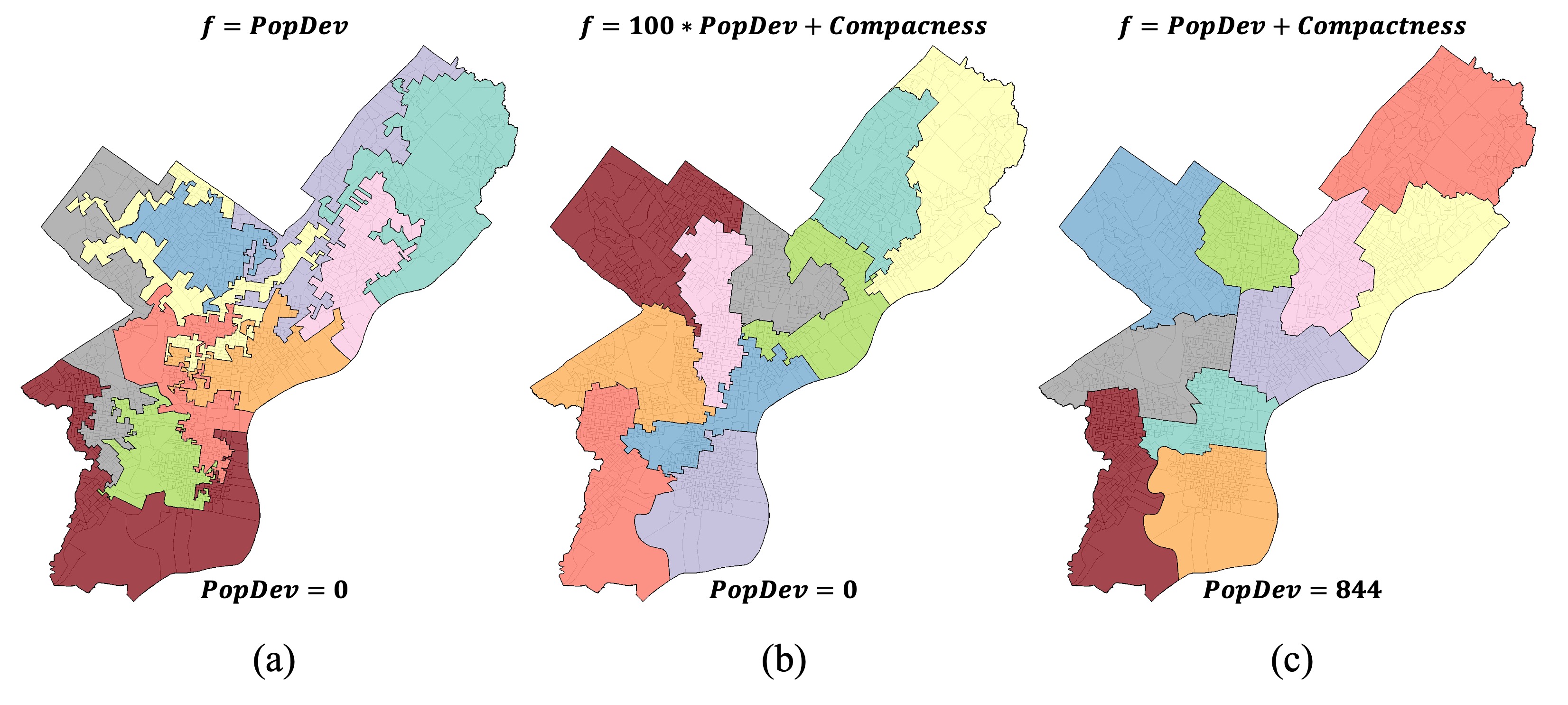}
    \caption{Selected Philadelphia City Council redistricting plans generated by \(\mathrm{Tabu}^*\) \emph{using 1,687 ward divisions}. (a) Optimizing population equality only under the contiguity constraint---global optimum attained (\(\mathit{PopDev} = 0\)). (b) Optimizing population equality and compactness with greater weight assigned to population equality---global optimum also attained. (c) Optimizing population equality and compactness with equal weights.}
    \label{fig:performance_philly}
\end{figure}

Figure~\ref{fig:performance_philly} shows representative plans produced by \(\mathrm{Tabu}^*\) under three configurations: (a) optimizing \(\mathit{PopDev}\) only; (b) optimizing \(\mathit{PopDev}\) and compactness with greater weight assigned to population equality; and (c) optimizing both criteria with equal weights. The plans in (a) and (b) both achieve the theoretical global optimum (\(\mathit{PopDev} = 0\)). Plan (c) maintains a small \(\mathit{PopDev}\) of 844 while achieving improved compactness. As noted in \citep{gopalanPhiladelphiaDistrictingContest2013} for this same Philadelphia data, it is a remarkable achievement to find a global optimum solution on population equality, let alone to consistently finding many different solutions of the global optimum.

\begin{figure}[!htbp]
    \centering
    \includegraphics[width=1\textwidth]{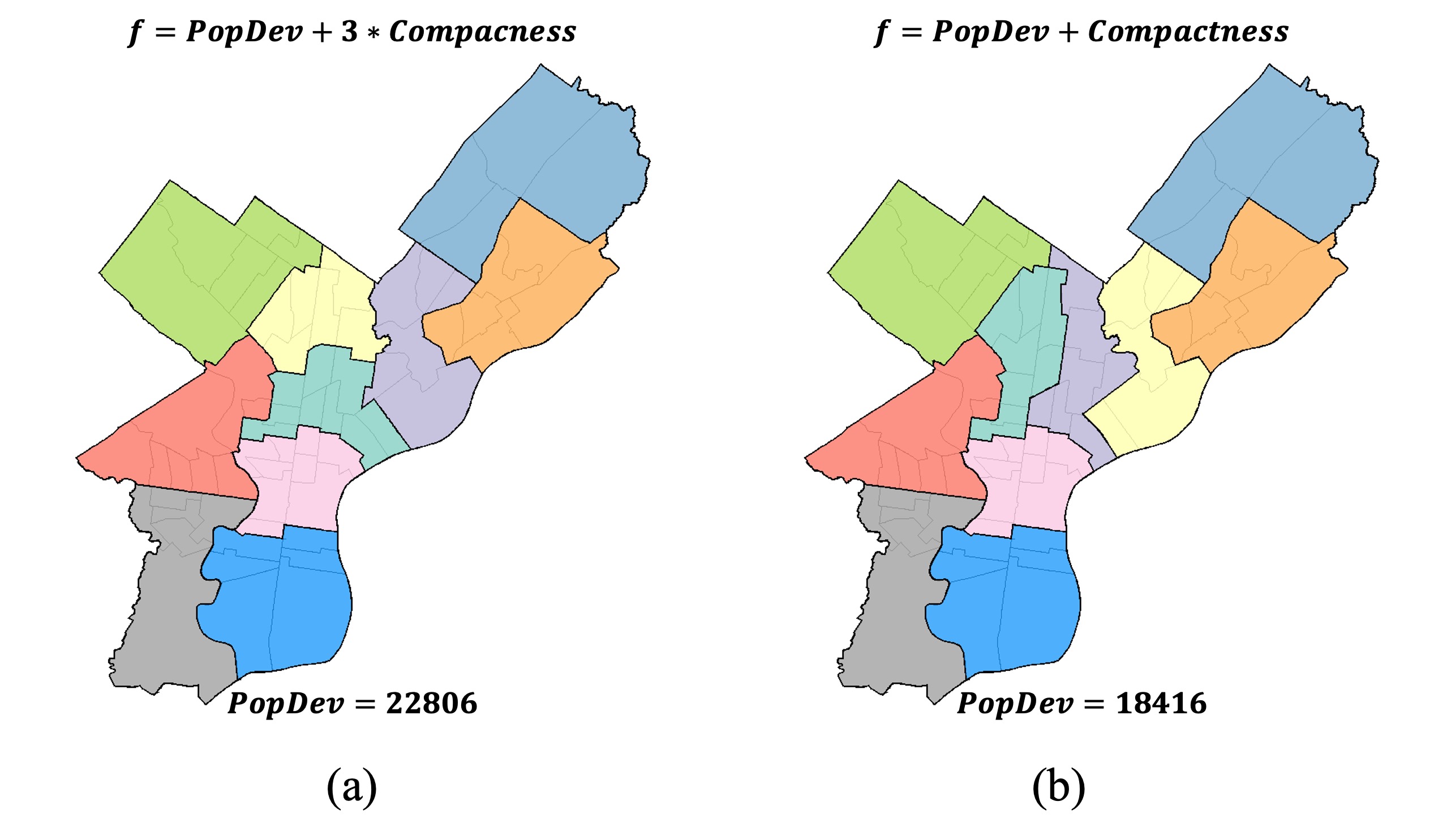}
    \caption{Two representative plans generated by \(\mathrm{Tabu}^*\) using \emph{66 wards} (instead of ward divisions) to preserve ward boundaries.}
    \label{fig:performance_philly_ward}
\end{figure}

If preserving ward boundaries is preferred, Figure~\ref{fig:performance_philly_ward} shows two plans generated by \(\mathrm{Tabu}^*\) using the 66 wards as the basic units, thereby fully preserving ward boundaries. As expected, this coarser aggregation reduces the flexibility for achieving perfect population equality; however, the total population deviation of both plans remains within 1.5 percent of the total population, which is generally acceptable in practice. The method can generate hundreds of plans within a minute; for clarity, we present only representative examples here.

In practice, the same framework can also incorporate additional criteria and interactive controls---such as drawing communities of interest, locking specific districts, or manually editing selected boundaries to incorporate local knowledge and considerations. However, this paper focuses on the core optimization algorithm and therefore does not present or evaluate these interactive features for human-computer collaboration.

Overall, these results indicate that, with its optimization power, the approach can optimize multi-criteria objectives to a fine level, allowing practitioners to focus more on interactive refinement to incorporate hard-to-quantify local knowledge and domain expertise, thereby enabling public participation and practical decision-support workflows.

\section{Conclusion and Discussion }

This paper presents an efficient and effective approach for strengthening the optimization power of combinatorial methods for practical spatial problems with hard contiguity constraints. Using redistricting as an illustrative application, we show that contiguity constraints can severely restrict the feasible neighborhood of trajectory-based search methods, limiting exploration and trapping searches in poor local optima. We address this bottleneck by introducing a contiguity-preserving neighborhood expansion based on composite moves and valid switches, which can be integrated into standard trajectory-based optimizers.

When combined with local greedy search, the Kernighan--Lin procedure, and Tabu search, the proposed approach substantially improves both solution quality and run-to-run reliability, with the Tabu-based integration, i.e., \emph{Composite-Move Tabu search} (CM-Tabu), achieving the strongest overall performance. Experiments on Iowa congressional redistricting demonstrate large and consistent gains in population-equality optimization, while the Philadelphia City Council case study illustrates effective multi-criteria optimization that balances population equality and compactness and supports interactive exploration of trade-offs. Across these studies, the approach can generate many diverse high-quality plans efficiently, making it well suited for real-world decision-support workflows.

Although this paper focuses on redistricting and trajectory-based optimization, the underlying framework is broadly applicable to other spatial combinatorial problems that require contiguity constraint and multi-criteria optimization, including school redistricting, location-allocation, traffic zone design, and service-territory planning. Adapting the framework to new applications may require problem-specific objective functions or constraints, but the core idea remains the same: explicitly leveraging contiguity structure to expand the feasible neighborhood search without violating hard constraints.

Several directions for future work follow naturally. First, the framework can be integrated with additional heuristic families, including population-based methods such as genetic algorithms, to further improve exploration and diversify solution sets. Second, richer practical constraints and preferences, including user-defined criteria and policy constraints, can be incorporated within the same move-evaluation framework to support interactive refinement. Overall, our results suggest that prioritizing optimization power within a flexible, constraint-respecting framework can help bridge the gap between theoretical algorithms and practical spatial decision-support systems.

\BlankLine
\BlankLine

\section*{Software Download}
The software package implementing the proposed approach, iRedistrict, will be made publicly available at \url{www.urbanxai.com}.

\clearpage

\bibliographystyle{elsarticle-harv}
\bibliography{_pub2026_GA_redistricting}

\end{document}